\pdfoutput=1
\documentclass[a4paper, 10pt, conference]{IEEEtran} 
\usepackage[left=0.75in,right=0.514in,top=1.458in,bottom=0.75in]{geometry}
\IEEEoverridecommandlockouts

\usepackage[utf8]{inputenc} % allow utf-8 input
\usepackage[T1]{fontenc}    % use 8-bit T1 fonts
\usepackage{hyperref}       % hyperlinks
\usepackage{url}            % simple URL typesetting
\usepackage{booktabs}       % professional-quality tables
\usepackage{amsfonts}       % blackboard math symbols
\usepackage{nicefrac}       % compact symbols for 1/2, etc.
\usepackage{microtype}      % microtypography
\usepackage{xcolor}         % colors
\usepackage{amsmath,amssymb,multicol,latexsym}
\usepackage{pgfplots,pgfplotstable}
\pgfplotsset{compat=1.16}
\usepackage{tikz}
\usepackage{hyperref}
\usepackage{cleveref}

\usepackage{todonotes}

\DeclareMathAlphabet{\mathcal}{OMS}{cmsy}{m}{n}

\usepackage{url}
\usepackage{graphicx}
\usepackage{cite}
\usepackage{flushend} % references even

\usepackage{tabu}
\usepackage{anysize} % Soporte para el comando \marginsize

\graphicspath{{figures/}}

\title{Self Supervised Clustering of Traffic Scenes using Graph Representations}

\author{
Maximilian~Zipfl$^{1}$,
Moritz~Jarosch$^{2}$, 
J.~Marius~Zöllner$^{1,2}$
\thanks{$^{1}$FZI Research Center for Information Technology, Karlsruhe, Germany
{\tt\small \{zipfl, zoellner\}@fzi.de}}%
\thanks{$^{2}$KIT - Karlsruhe Institute of Technology, Karlsruhe, Germany
{\tt\small urkqe@student.kit.edu, marius.zoellner@kit.edu}}%
}

\date{June 2022}

\begin{document}

\maketitle

\begin{abstract}
Examining graphs for similarity is a well-known challenge, but one that is mandatory for grouping graphs together. 
We present a data-driven method to cluster traffic scenes that is self-supervised, i.e. without manual labelling.
We leverage the semantic scene graph model to create a generic graph embedding of the traffic scene, which is then mapped to a low-dimensional embedding space using a Siamese network, in which clustering is performed. 
In the training process of our novel approach, we augment existing traffic scenes in the Cartesian space to generate positive similarity samples.
This allows us to overcome the challenge of reconstructing a graph and at the same time obtain a representation to describe the similarity of traffic scenes.
We could show, that the resulting clusters possess common semantic characteristics.
The approach was evaluated on the INTERACTION dataset.
\end{abstract}

% ==============
%  Introduction
% ==============

\section{Intruduction}
Scenario-based testing of highly automated driving functions has become increasingly important in recent years. This technique has the advantage over conventional, statistical validation methods that important scenarios can be tested in a targeted manner without having to drive huge amounts of kilometres \cite{shalev-shwartz_formal_2018}.  
Since it is infeasible to test all possible scenarios, they are systematically generated either by a knowledge-based approach, which uses expert knowledge, guidelines or traffic rules, or by data-driven approaches.
In the latter case, scenarios must be classified and grouped beforehand in order to obtain a high degree of coverage of all possible scenarios with just a few concrete samples. To avoid manual labelling, the clustering should happen automatically.
Once scene-clusters have been formed, they can be classified and analysed. Subsequently, exemplary scenarios can be reconstructed in a simulation, for example, to validate the highly automated driving function.

The most important components for clustering traffic scenarios is the description and their comparability \cite{bagschik_ontology_2018}.
In this work, we want to address the latter and propose a Graph Neural Network (GNN) based approach that allows to cluster traffic scenes described by scene graphs without prior manual labelling.
To minimize the complexity level of the task, only traffic scenes, i.e. snapshots at discrete time steps, are considered. The temporal information could be added later in a downstream step. 

Nevertheless, we need an approach that transforms the traffic scene graphs into a representation, where the similarity of two scenes can be described explicitly.
Different architectures are possible to create an encoding from graphs to be used for clustering. One option are Graph (Variational) Autoencoders. (Variational) Autoencoders are able to encode deep semantic understanding into latent spaces, shown by works like Cristovao et al. \cite{cristovao_generating_2020}.
But applying them to graphs remains notoriously difficult. Especially for graphs with rich edge features and/or multigraphs (graphs that can have multiple edges between the same nodes), both properties that the graphs in our application possess.
\begin{figure}[t]
    \centering
    \includegraphics[width=0.95\linewidth]{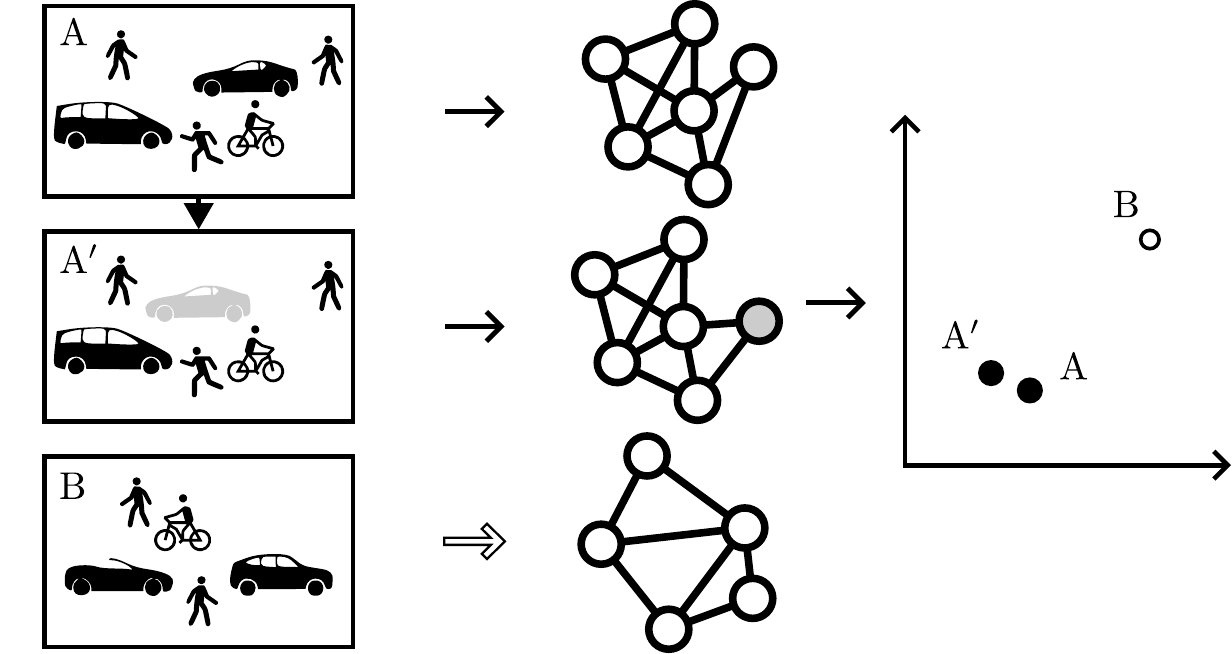}
    \caption{Schematic illustration of our approach: Creating a sample triplet for self supervised traffic scene clustering}
    \label{fig:approach_topright}
    \vspace{-2ex}
\end{figure}
Another great challenge is the graph generation or decoder part. 
Generating graphs from a fixed size encoding vector can be achieved in an all-at-once fashion or recurrently. Both variants come with their own drawbacks \cite{simonovsky_graphvae_2018}, like limiting the type of graph, its size or connectivity. A second problem is the comparability of graphs. To train such an Autoencoder a metric for comparing them is needed.
Therefore, one has to overcome the graph isomorphism problem \cite{toran_hardness_nodate}. This addresses the question if two graphs are isomorphic and by extension to find a matching that compares nodes and edges to each other in both graphs. From there, a loss metric could be easily calculated.
To this date, the graph isomorphism problem is not known to be solvable in polynomial time. Adding to this limitation, finding a sufficient matching on graphs that are multigraphs or have multidimensional edge features makes it even more unfeasible. Simonovsky et al. \cite{simonovsky_graphvae_2018} implemented such an approximate graph matching and used it on Autoencoders, but only small graphs, up to 38 nodes and only with singular connectivity between nodes, could be generated. Other approaches \cite{kipf_variational_2016} simply expect the network to learn to produce the correct matching implicitly.
This can be a feasible approach, considering that Xu et al. \cite{xu_how_2019} showed that their GNN is as powerful as the popular Weisfeiler-Lehman graph isomorphism test.

Another idea is to construct a data driven encoder that can distinguish and measure the difference between graphs.
This encoder could also be used to identify graph classes by identifying clusters in the encoding space.

This work implements and analyses such an encoder, utilizing Siamese networks and contrastive learning to facilitate comparing traffic scenes (see \Cref{fig:approach_topright}).

In order to process traffic scenes with a machine learning method, they have to be converted into a machine-readable format. For this purpose, we leverage the semantic scene graph presented in \cite{zip_towards_nodate}. Thereby, absolute positions of the traffic participants are neglected strategically, and focus is put on interactions between entities. Traffic participants are represented as nodes in the graph. Depending on the road network topology and the relative location of the traffic participants, heterogeneous edges are created, that encode relations between the traffic participants. There are three different edge types: longitudinal - when traffic participants are on the same lane, lateral - on two parallel lanes, or intersecting - when two lanes intersect or merge. Velocity information and the classification of traffic participants is stored in the node attributes. The information about the edge type and the distance in Frenet space along the lanes is stored in the edge attributes.
This abstraction allows traffic scenes to be described independently of the road geometry and improves the comparability.
For the detailed implementation and more background information regarding the scene graph, please refer to \cite{zip_towards_nodate}.

This paper is structured as follows: In \Cref{sec:related_work} we provide a review of existing approaches for graph representations in machine learning and other related work. In \Cref{sec:methodology} we give a detailed insight to the challenge and the methodology of our approach. Afterwards, in \Cref{sec:experiments} we apply our approach on a traffic motion dataset. Finally, in \Cref{sec:conclusion} we conclude this contribution.

% ==============
%  Related Work
% ==============
\section{Related Work}
\label{sec:related_work}

\subsection{Self-supervised Contrastive Learning}
The basic idea of contrastive learning is that the representations of the same similar, \emph{positive} samples in the latent state space are contracted, while \emph{negative} samples are shifted away from the initial sample (\emph{anchor}). 
In many application areas, often no label of the data is available. A positive sample can then be generated by data augmentation. Negative samples can be generated by randomly selecting from the remaining data. 
In recent years, contrastive learning, i.e. the combination of negative and positive samples, has been shown to improve the performance of self-supervised representation \cite{wu_unsupervised_2018, khosla_supervised_2021, sohn_improved_nodate, doersch_unsupervised_2016}. 
Here, the work of Ma et al. \cite{ma_multi-agent_2021} is also worth mentioning, which uses trajectory data of traffic participants to cluster traffic scenarios in a self-supervised manner.

Recent research has focused not only on contrastive machine learning approaches for structured input data, but also for unstructured data such as graphs. 

\subsection{Graph Contrastive Learning}
With graphs being discrete structures, applying machine learning and especially neural networks to them requires a specific framework.
Approaches to such Graph Neural Networks (GNNs) are gaining more widespread traction in recent years \cite{wu_comprehensive_2021}. In this paper, we will focus on so-called Message Passing networks, that leverage the existing graph structure and that can handle various graph types well.

We will use this scheme to perform self-supervised representation learning on whole graphs. To be more precise, we will conduct contrastive learning in order to obtain encodings that can be used to compare and cluster traffic scenes.

Zhu et al. \cite{zhu_survey_2022} provide a great overview over different contrastive learning approaches on graphs and the design considerations and dimensions, that include Data Augmentation, Contrasting Modes, Contrastive, Objectives and Negative Mining Strategies.

To the best of our knowledge, only a few papers have been published on this topic of contrastive learning to embed whole graphs \cite{you_graph_2020, hassani_contrastive_2020, sun_infograph_2020}.

InfoGraph \cite{sun_infograph_2020} for example doesn't augment the graphs, but works by discriminating representations of a whole graph with the representations of single nodes of other graphs. They aim to learn aspects of the data that are shared within substructures and aim to maximize mutual information between whole graphs and nodes. This also means that this approach doesn't use graph data augmentations in learning.

Another approach, using data augmentation, is presented by Hassani et al. \cite{hassani_contrastive_2020}. They augment graphs using diffusion views, but are not augmenting initial node features. The same subset of nodes from both graphs are sampled, comparing it to cropping in the visual domain. They are also using two independent GNNs, but shared Multilayer Perceptrons (MLPs) for creating the node and graph encodings. 

Our encoder architecture is similar to the one used by You et al. \cite{you_graph_2020}, consisting of GNN layers, followed by a readout function and a final MLP. They studied the effect different augmentations, which were applied directly on the graph level, have on contrastive graph learning. The augmentations they studied were: node dropping, edge perturbations, attribute masking and subgraphs.

In this work we propose a novel augmentation scheme, that modifies both the node and edge features, as well as the graph topology in a significant, yet semantically coherent manner. 
With our approach, only valid scenes are generated by incorporating the street layout into the augmentation.

% =====================
%  Problem Formulation
% =====================
% \section{Problem Formulation}
% \label{sec:problem_formulation}

% =============
%  Methodology
% =============
\section{Methodology}
\label{sec:methodology}
Different to related work, our approach does not start with graphs as input, but rather with the generation of these graphs.
The decisive step is that the traffic scene data is augmented directly on object list level, thus augmented in the Cartesian space.
Subsequently, a graph triplet can be formed, whereby pseudo labels are generated for all three graphs in order to be able to generate a loss. 

\subsection{Dataset, Augmentation and Graphgeneration}
The starting point of our processing pipeline are traffic scenes of a motion dataset.
Exemplarily, we use the INTERACTION dataset \cite{zhan_interaction_2019} to demonstrate the functionality of our approach. This dataset focuses on the behaviour of vehicles on roads in different countries.
Since the dataset was recorded by a drone, the same location, in this case intersections or roundabouts, is observed in each sequence. In addition to the object tracks, the corresponding roads are provided as HD maps. Each object in the dataset is defined by its pose in Cartesian space, classification and velocity at each time step sampled at 10 Hz.

In our approach, we modify the traffic scene directly
on the object list, rather than changing the graph features directly. This has the advantage that realistic and consistent traffic scenes continue to result. 
Randomly adding, deleting or modifying edges and nodes in the graph representation can result in scenes that cannot occur in reality. 

The position and velocity of individual traffic participants are randomly modified.
Traffic participants of a scene whose entity state $\mathcal{X}^i$ is to be modified are selected with the probability $p_{entity}$.
The variation of the position $(x,y)$ is described by normal distributions $N_{x} \sim (x,\sigma_{pos})$, $N_{y} \sim (y,\sigma_{pos})$  and the variation of the velocity $(\dot{x}, \dot{y})$ by $N_{\dot{x}} \sim (x,\sigma_{pos}), N_{\dot{y}} \sim (y,\sigma_{pos})$.

\begin{figure}
    \centering
    \def\svgwidth{0.80\columnwidth}
    %% Creator: Inkscape 1.2 (1:1.2+202205231507+da316b6974), www.inkscape.org
%% PDF/EPS/PS + LaTeX output extension by Johan Engelen, 2010
%% Accompanies image file 'augmented_scenes_example.pdf' (pdf, eps, ps)
%%
%% To include the image in your LaTeX document, write
%%   \input{<filename>.pdf_tex}
%%  instead of
%%   \includegraphics{<filename>.pdf}
%% To scale the image, write
%%   \def\svgwidth{<desired width>}
%%   \input{<filename>.pdf_tex}
%%  instead of
%%   \includegraphics[width=<desired width>]{<filename>.pdf}
%%
%% Images with a different path to the parent latex file can
%% be accessed with the `import' package (which may need to be
%% installed) using
%%   \usepackage{import}
%% in the preamble, and then including the image with
%%   \import{<path to file>}{<filename>.pdf_tex}
%% Alternatively, one can specify
%%   \graphicspath{{<path to file>/}}
%% 
%% For more information, please see info/svg-inkscape on CTAN:
%%   http://tug.ctan.org/tex-archive/info/svg-inkscape
%%
\begingroup%
  \makeatletter%
  \providecommand\color[2][]{%
    \errmessage{(Inkscape) Color is used for the text in Inkscape, but the package 'color.sty' is not loaded}%
    \renewcommand\color[2][]{}%
  }%
  \providecommand\transparent[1]{%
    \errmessage{(Inkscape) Transparency is used (non-zero) for the text in Inkscape, but the package 'transparent.sty' is not loaded}%
    \renewcommand\transparent[1]{}%
  }%
  \providecommand\rotatebox[2]{#2}%
  \newcommand*\fsize{\dimexpr\f@size pt\relax}%
  \newcommand*\lineheight[1]{\fontsize{\fsize}{#1\fsize}\selectfont}%
  \ifx\svgwidth\undefined%
    \setlength{\unitlength}{458.2616755bp}%
    \ifx\svgscale\undefined%
      \relax%
    \else%
      \setlength{\unitlength}{\unitlength * \real{\svgscale}}%
    \fi%
  \else%
    \setlength{\unitlength}{\svgwidth}%
  \fi%
  \global\let\svgwidth\undefined%
  \global\let\svgscale\undefined%
  \makeatother%
  \begin{picture}(1,0.9371182)%
    \lineheight{1}%
    \setlength\tabcolsep{0pt}%
    \put(0,0){\includegraphics[width=\unitlength,page=1]{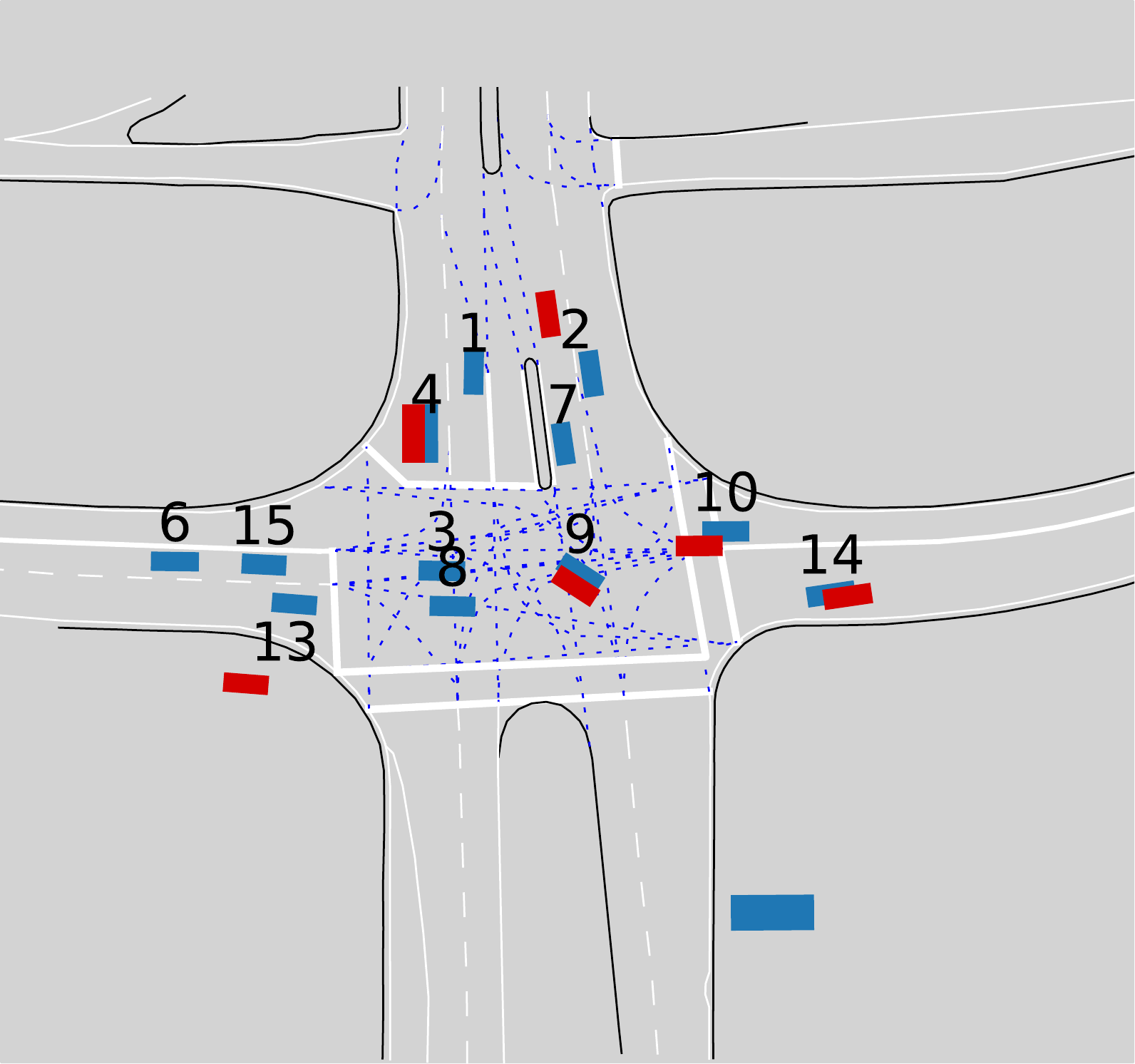}}%
    \put(0.75058868,0.12082026){\color[rgb]{0,0,0}\makebox(0,0)[lt]{\lineheight{1.25}\smash{\begin{tabular}[t]{l}original\end{tabular}}}}%
    \put(0.75037551,0.04057532){\color[rgb]{0,0,0}\makebox(0,0)[lt]{\lineheight{1.25}\smash{\begin{tabular}[t]{l}augmented\end{tabular}}}}%
    \put(0,0){\includegraphics[width=\unitlength,page=2]{augmented_scenes_example.pdf}}%
  \end{picture}%
\endgroup%

    \caption{Different position of traffic participants between the original and the augmented traffic scene}
    \label{fig:differences_aug_scene}
\end{figure}

This augmentation step will then be repeated for the whole dataset to generate more positive samples. In the process, an attempt is made to avoid collisions between traffic participants.
In \Cref{fig:differences_aug_scene} the differences between an exemplary, original traffic scene and its augmented counterpart are depicted. Blue rectangles describe the original poses of traffic participants, red rectangles display the augmented poses. 
Most positions are changed only minimally (see 9, 10, 14). However, due to the random shift, traffic participants can also change lanes, causing vehicle 2 to drive in front of vehicle 7 instead of next to it after augmentation. Furthermore, traffic participants may be moved outside the road geometry (see vehicle 13). As a result, this vehicle is not considered further.

Subsequently, both the modified and the raw traffic scenes are converted into corresponding scene graphs $G_{+} amd G_0$, which serve as the input format for the machine learning process.
Each graph $G = (V,E)$ is defined by the nodes $v \in V$ which represent the traffic participants and the edges $e_{ij}$ between two nodes ($v_i,v_j$), which represent their relations. 
It may happen that traffic scenes with few vehicles do not have edges in the graph representation. This is the case, for example, when two vehicles are driving on oncoming lanes that do not cross each other, or when one vehicle is driving by itself on the intersection.
Such graphs without edges are not considered in our framework, because the information propagation between nodes plays a crucial role in our approach. 
However, this is not a limitation of the presented approach, but explicitly filters trivial correlations.

In summary, the scenes are augmented specifically on the basis of a pattern and not purely randomly. The velocities and positions are reasonably adjusted based on the initial scene. By incorporating the scene graph model and the road layout, disproportionate modifications and the resulting structure of the graphs can be limited to realistic examples.

\subsection{Graph Representation and Learning Approach}

\begin{figure}[htbp]
    \centering
    \includegraphics[width=1.0\linewidth]{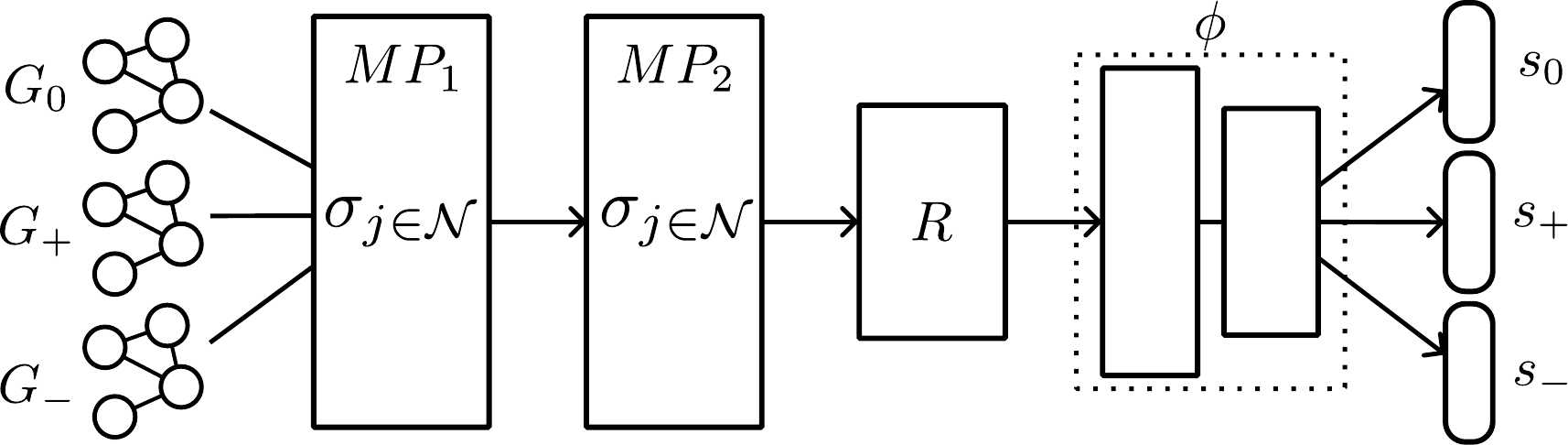}
    \caption{Representation pipeline, which maps graphs onto the desired feature space.}
    \label{fig:representation_pipeline}
\end{figure}

Each triplet of a traffic scene, consisting of the anchor $G_{0}$, the positive graph sample $G_+$ and the negative graph sample $G_-$ should be mapped onto the embedding space $S = (S_x,S_y) \subseteq \mathbb{R}$. This is done by the graph representation pipeline depicted in \Cref{fig:representation_pipeline}. The nodes' neighborhood information is captured by two successive message passing operations $MP_{k}$.

The concept of message passing can be understood with the following formula:
 \begin{align}
     v_i^k = \gamma^k(v_i^{k-1}, \sigma_{j \in N(i)}(m^k(v_i^{k-1}, v_j^{k-1}, e_{j,i}))).
     \label{eq:message_passing}
 \end{align}

 In each network layer $k \in \{1,..,K\}$, the state of the $i$-th node $v_i^k$ is updated by firstly calculating a message along all incoming edges to the node. This is done by applying the message function $m^k$ to the attributes of the nodes $v_i^{k-1}, v_j^{k-1}$ and the edges $e_{i,j}$ connecting them. Messages are aggregated using an aggregation function $\sigma$. Often times $min$, $max$, $sum$ or $mean$ are used. The resulting message is then passed through a final update function $\gamma$, alongside the original node attributes of the node's state $v_i^{k-1}$, resulting in the final new state $v_i^k$. 
 Appropriate functions for $m$ and $u$ are parametrized through neural network layers.

The resulting node states $v^K$ are merged by a readout layer $R$. This function can be a $mean$, $min$, $max$, or $sum$ operation, but more complex methods are also possible \cite{ying_hierarchical_2018}.

The resulting latent feature vector of the graph is then mapped onto the final feature space by an MLP $\phi$ with an intermediary activation function.

Learning is conducted using the triplet loss function $\mathcal{L}$ from \Cref{eq:triplet}.
\begin{align}
    \label{eq:triplet}
    \mathcal{L}(s_0, s_+, s_-) = max\big(d(s_0, s_+) - d(s_0, s_-) + M, 0\big)
\end{align}

With $s_0$, $s_+$ and $s_-$ being graph encodings ($s_0, s_+, s_- \in S$). In our case, $s_0$ is the encoding of the original graph, the so-called anchor. The augmented encoding, the positive sample, is $s_+$ and a sampled, thus differing graph results in the negative example $s_-$. Sampling negatives is a broad research area itself \cite{zhu_survey_2022}. For our use case, negative samples are sampled at uniform from the rest of the batch. The function $d$ can be any suitable distance metric, like euclidean distance or cosine similarity. The margin $M$ is used to distance negative samples by a desired spread.

% ============
%  Evaluation
% ============
\section{Experiments}
\label{sec:experiments}

\subsection{Training Setup}

As shown in \Cref{fig:representation_pipeline}, the Siamese network architecture consists of two GNN layers $MP_1$, $MP_2$, followed by a readout function $R$ and an MLP $\phi$ with one hidden layer.

Both GNN layers use a single linear layer for the message function $m^k$ and the update function $\gamma^k$. The outputs of those layers are each passed through a LeakyRelu activation function. $MP_1$ uses both, node (6 dimensions) and edge features (7 dimensions) in the message function, while $MP_2$ uses only the node features (6 dimensions) that are the result of the first GNN layer. As aggregation function $\sigma$ the $sum$ is used.

As readout function $R$,  the $mean$ is used over all node features of a graph.

The resulting encoding (6 dimensions) is passed through $\phi$, with a 6 dimensional hidden layer, obtaining a 2 dimensional encoding $(S_x, S_y)$ in the embedding space $S$ that is used for training and subsequent experiments. The activation function for the hidden dimensions is LeakyRelu and the final output is processed through a $tanh$ function.

The dataset used for training is the aforementioned INTERACTION dataset \cite{zhan_interaction_2019}. Training was conducted on 2842 scenes of one roundabout.
The resulting graphs have 6 node features, which are velocity, yaw of the vehicle and one hot encodings of the vehicle type (4 dimensions).

And 7 edge features consists of a one hot encoding of interaction type (3 dimensions), the distance along Frenet path, the centerline distance root node, the centerline distance target node and the probability of the edge.

For more detailed information on the attributes, please refer to the scene graph paper \cite{zip_towards_nodate}.

The aforementioned triplet loss $\mathcal{L}$ in \Cref{eq:triplet} is used in conjunction with the Adam optimizer. A learning rate of $0.001$ is used, and the triplet margin is set to $0.5$. The distance function used in the triplet loss is the euclidean distance.

Training is conducted in $1400$ epochs, with batches of size $533$. The model is trained on $75\%$ of the traffic scenes and validated using the rest.
Negative sampling was conducted uniformly over the batch, excluding the positive sample.

For generation of the augmented data, we use $\sigma_{pos} = 1.5\,m$ and $\sigma_{vel}=2.0 \frac{m}{s}$ with $p_{entity}=0.5$.

\subsection{In-Depth Analysis}
\subsubsection{Encoding Space}
Viewing the encoded traffic scenes in a scatter plot already makes it obvious that clusters have formed. Coloring the encodings using statistics of the original graphs result in interesting views on the formed clusters.

\begin{figure}[tbp]
    \centering
    \includegraphics[width=0.85\linewidth]{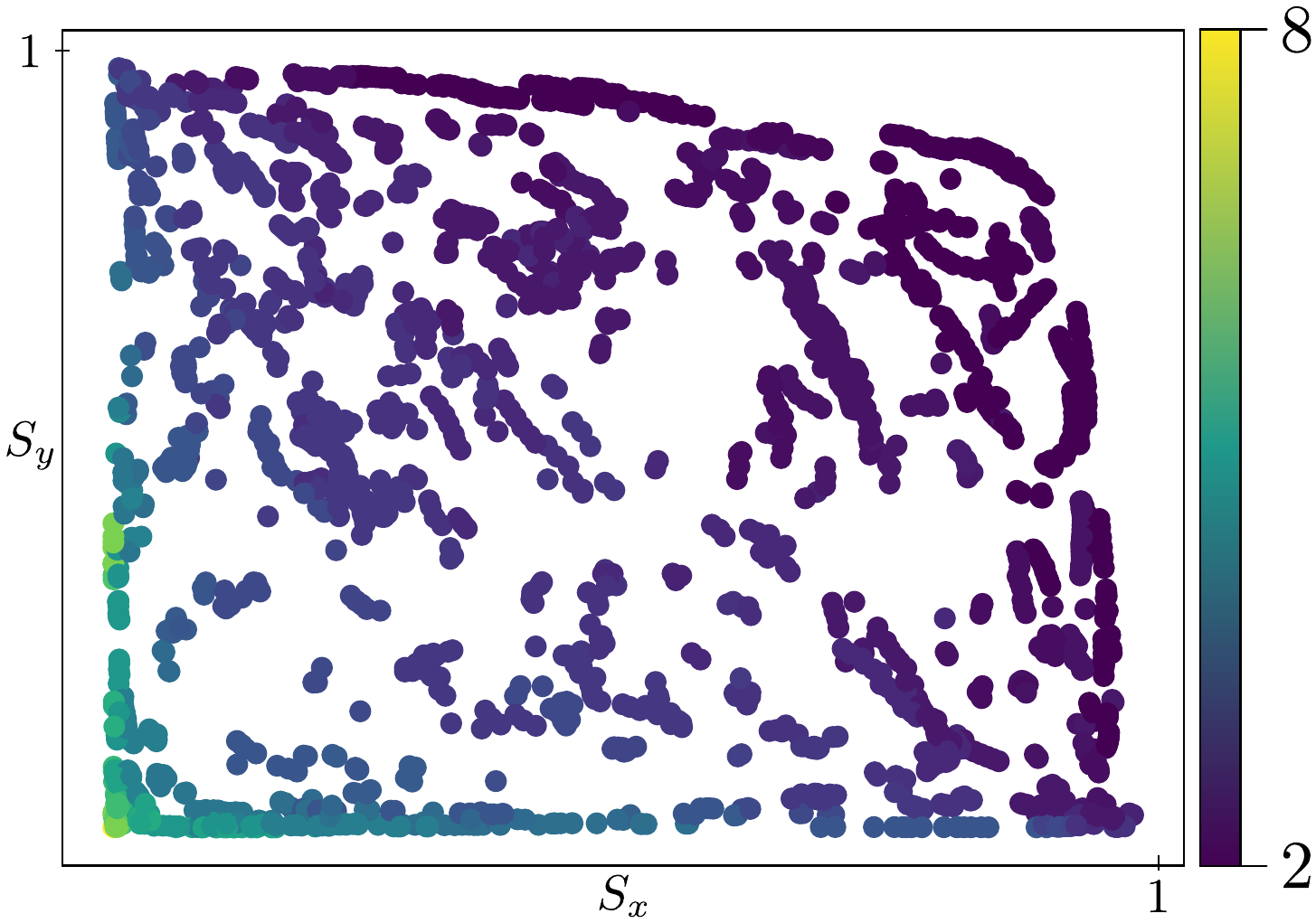}
    \caption{Encoded traffic scenes, mapped to the encoding space $S$, colored according to the average number of cars in each scene.}
    \label{fig:cars}
\end{figure}

Encoding the number of cars in each scene in the same fashion reveals (\Cref{fig:cars}) a gradient showing, that scenes with many cars tend to be clustered in the lower left, while scenes with fewer cars tend to cluster in the upper right. 
This shows that the number of vehicles for the network is a decisive criterion for the differentiation of traffic scenes. This can also be seen in \Cref{fig:scene_consec}. For example, a traffic jam scene ultimately requires several vehicles.

This shows that at least some meaningful structure is formed on a global level. 

\subsubsection{Clustering Evaluation}
\begin{figure}[tbp]
    \centering
    \includegraphics[width=0.85\linewidth]{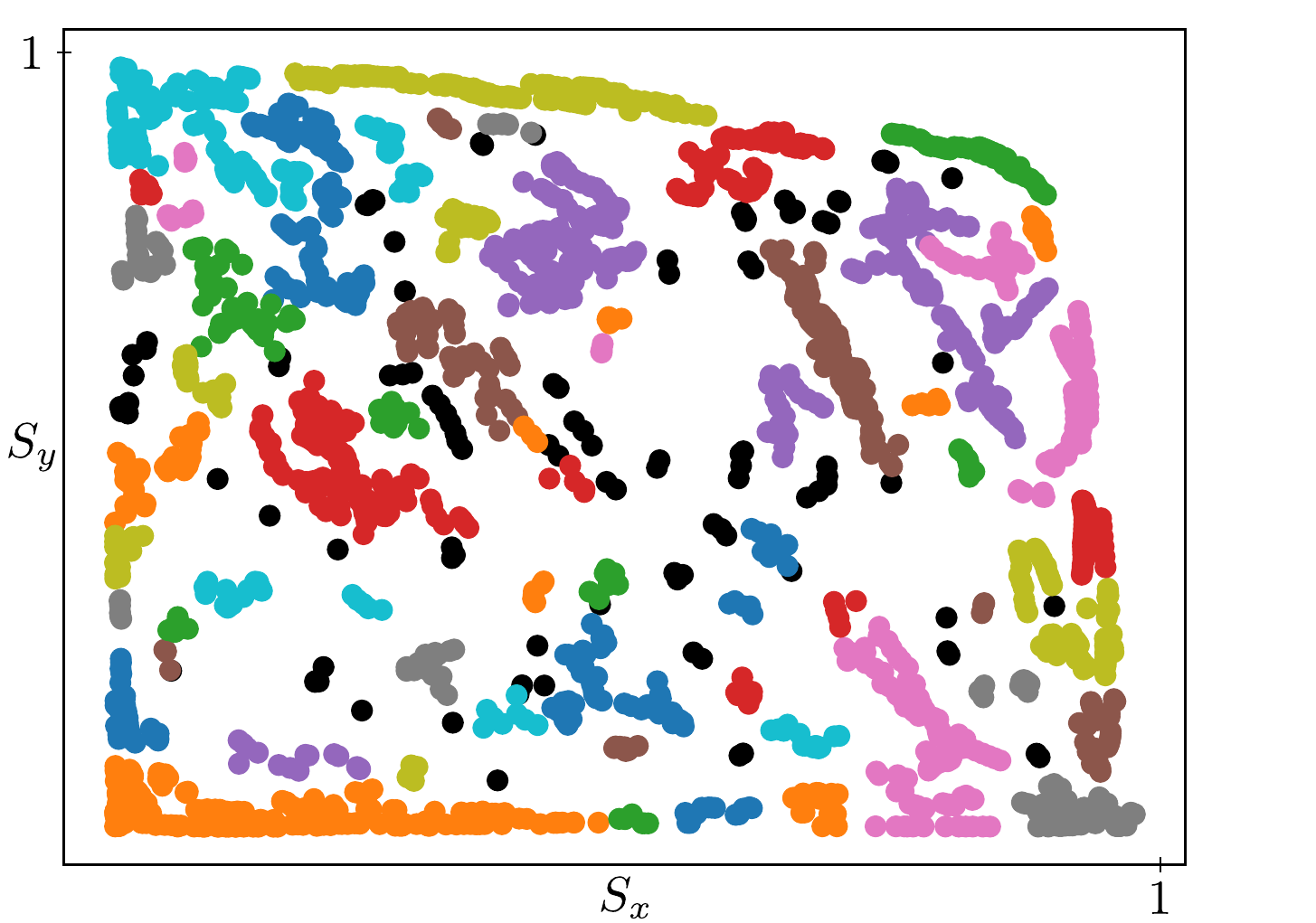}
    \caption{Each dot represents a graphic scene embedding in the two-dimensional encoding space $S$. Samples in one cluster are marked in the same color}
    \label{fig:clusters}
\end{figure}

To further analyze the formed clusters on their semantic meaning, the DBSCAN clustering algorithm ($\epsilon=0.05$, $samples_{min}~=~5$) was used on the encodings. The parameters are found by using the elbow method \cite{noauthor_elbow_2022}. This resulted in 75 distinct clusters and outliers (black), shown in \Cref{fig:clusters}.

For the qualitative analysis of the approach, two traffic scenes from each of two different clusters are shown in the following figures (\Cref{fig:scene_consec}, \Cref{fig:scene_parking_wait}).
In general, a cluster contains traffic scenes that are successive in time, as was to be expected. This shows that almost identical scenes are grouped together. In addition, scenes that were recorded at a different time but have a similar vehicle constellation are also assigned to the same cluster. For example, in \Cref{fig:scene_consec} one can observe several vehicles driving in a chain behind each other. Nevertheless, there are also areas that differ, yet in this cluster this one feature is represented in every sample.
This scheme is present in most of the clusters. The example in \Cref{fig:scene_parking_wait} is worth highlighting, here the network has automatically grouped traffic scenes that include the parking vehicle (ID: 93), although the velocity of each participant is included, but not the information of the classification "parking".

Nevertheless, some traffic scenes have been turned into a cluster, where one can't identify a recognizable pattern when you look at the original scene. This is due to the fact that the network was not given any classification information to group traffic scenes into clusters. This only means that humans cannot classify these clusters. Nonetheless, similarities have been found here, which may also have their own justification of existence, but which cannot be classified at first glance by common sense. 

\begin{figure}[htbp]
    \centering
    {{\includegraphics[width=0.47\columnwidth]{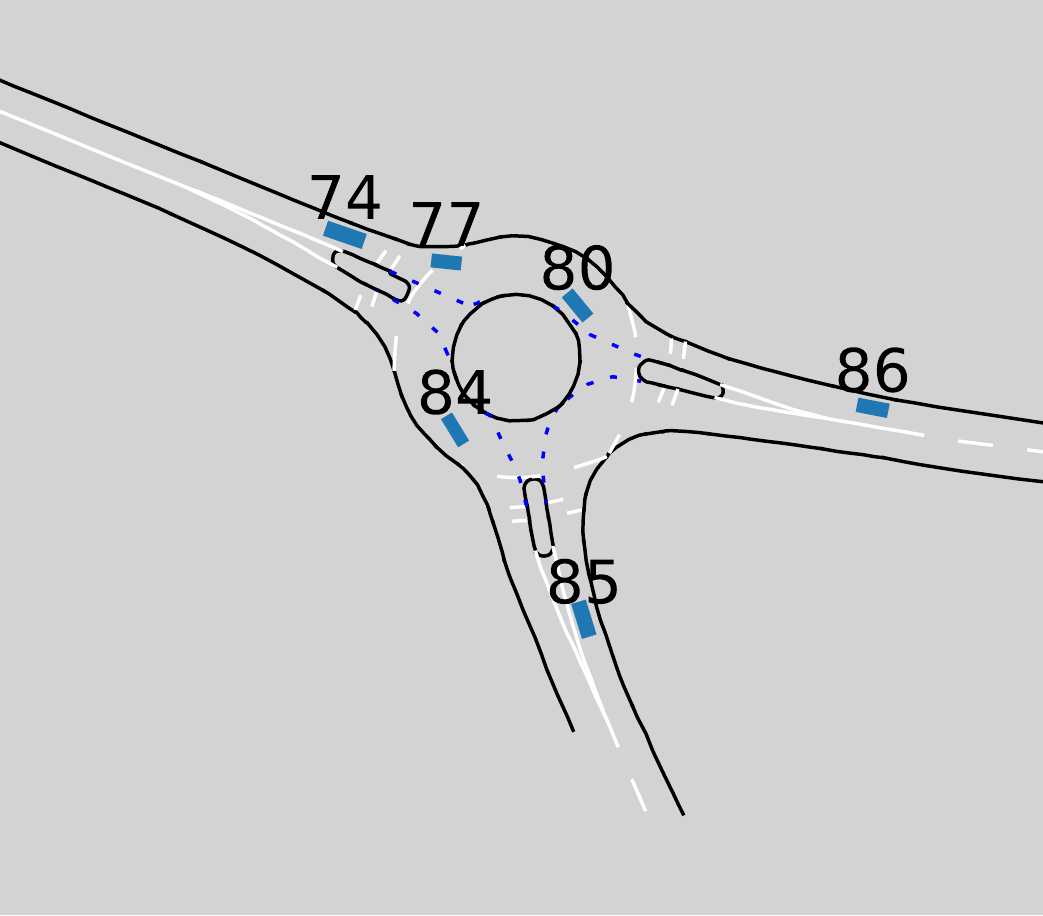} }}%
    {{\includegraphics[width=0.47\columnwidth]{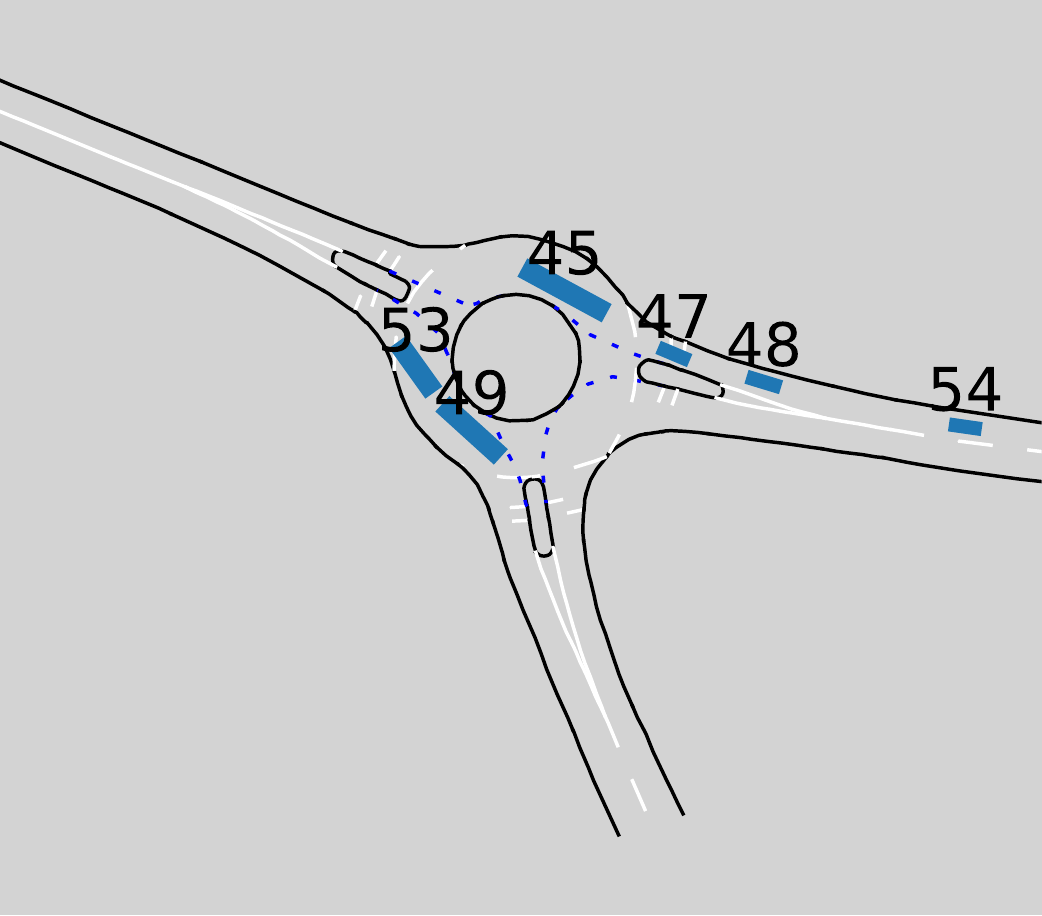} }}%
    \caption{Two exemplary traffic scenes of the same cluster, that has the characteristic that several vehicles drive behind each other. Vehicles are described by blue rectangles, road boundaries are indicated with black and white lines.}
    \label{fig:scene_consec}
\end{figure}

\begin{figure}[htbp]
    \centering
    {{\includegraphics[width=0.464\columnwidth]{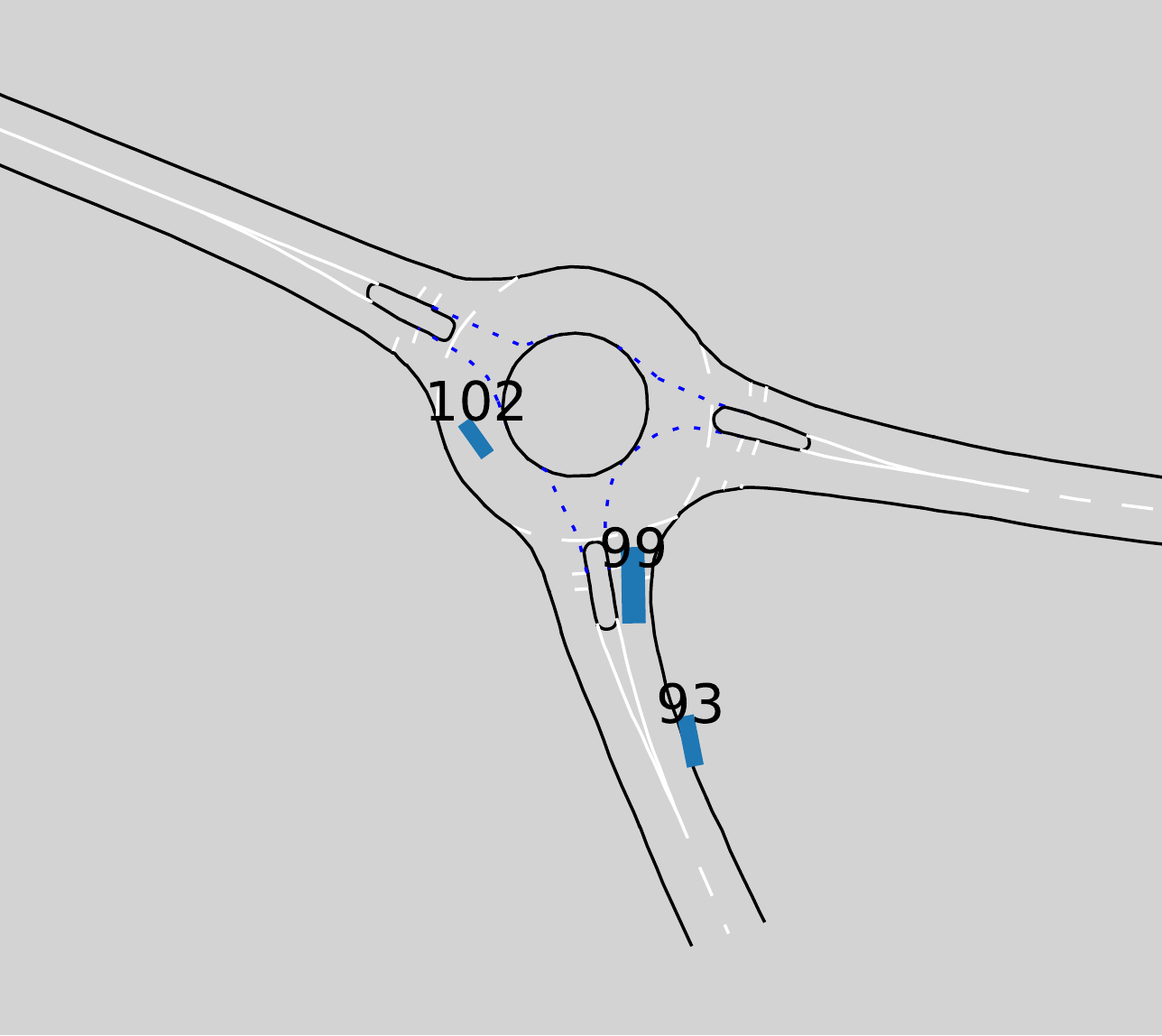} }}%
    {{\includegraphics[width=0.47\columnwidth]{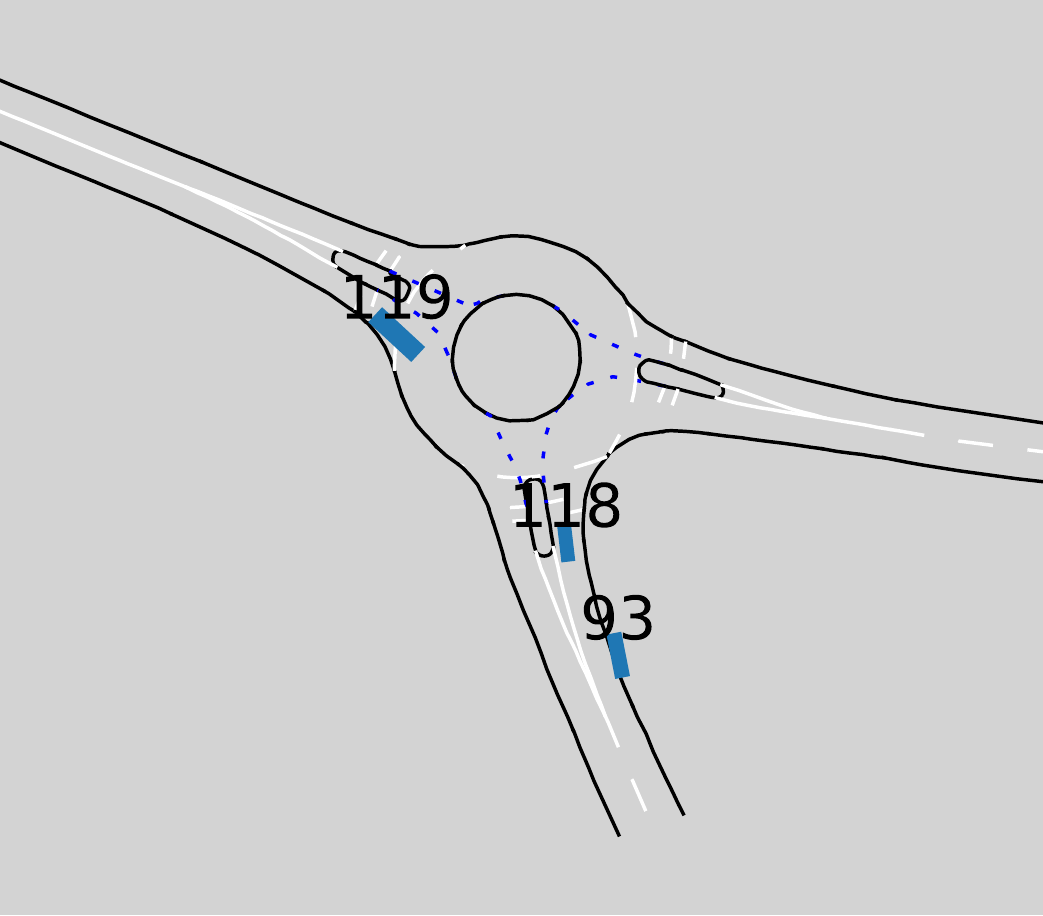} }}%
    \caption{Two exemplary traffic scenes of the same cluster, that has the characteristic that several vehicles are in the roundabout and there exist a parking vehicle. Vehicles are described by blue rectangles, road boundaries are indicated with black and white lines.}
    \label{fig:scene_parking_wait}
\end{figure}
\subsubsection{Attribute Representation}
\begin{figure}[tbp]
    \centering
    \includegraphics[width=0.85\linewidth]{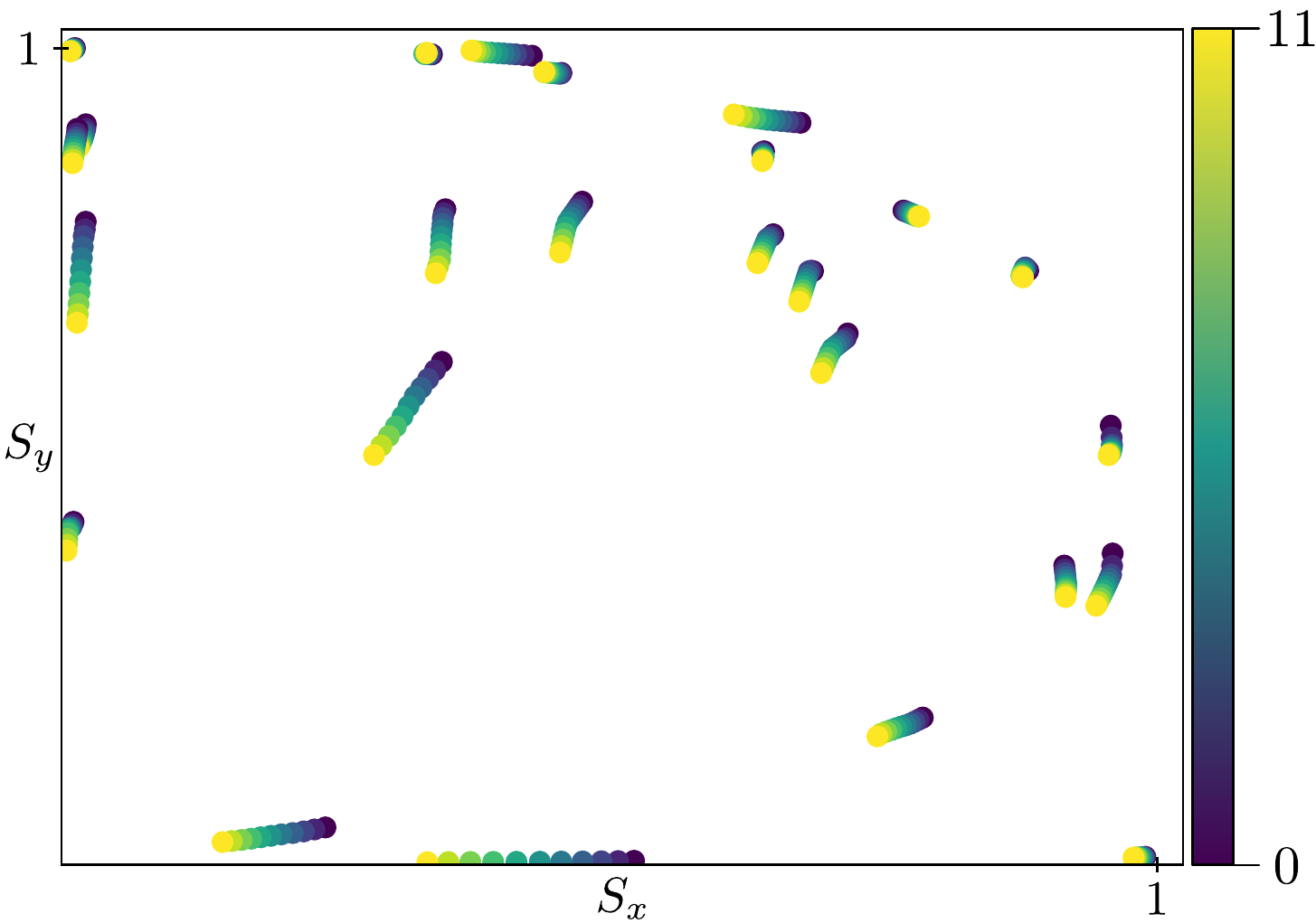}
    \caption{
    Each dot represents a graphic scene embedding in the two-dimensional encoding space $S$ with increasing (dark blue $\rightarrow{}$ yellow) velocities for each traffic participant. Dark blue dots represent the original scene with no velocity modifications.}
    \label{fig:altered}
\end{figure}

Lastly, we visualize the impact that changing singular attributes in the original graph has on the encoding. For better clarity, we select 25 random traffic scenes in different clusters and increment the velocity of all traffic participants by $0.5 \frac{m}{s}$ ten times, resulting in eleven various scenes in each case. All samples and their altered versions are encoded in $S$ and plotted in \Cref{fig:altered}.
Interestingly, many samples are pulled towards the lower left, the faster they become. 
The further an altered data point is shifted, the more this traffic scene differs from the initial scene. This can be seen especially in traffic scenes with many traffic participants driving close behind each other at the same time (see \Cref{fig:altered} bottom left and \Cref{fig:scene_consec}).

% ============
%  Conclusion
% ============
\section{Conclusion and Outlook}
\label{sec:conclusion}

In this work, we introduce a novel approach for augmenting graph data for contrastive learning in suitable datasets, and successfully apply it to traffic scene graphs.
We showed that our approach, even with a simple network architecture, is able to find similarities in traffic scenes, that can be used to create clusters in semantically meaningful ways.

Due to the ability of the network to expand freely, many clusters with recognizable characteristics are grouped together, but some groups have also formed, that do not make it possible to make a concrete, human understandable statement about the type of scene. 

In the future, we aim to explore this topic even further, for example by enlarging the dataset with scenes from other intersections, employing more sophisticated negative sampling methods and by searching for more refined network architectures.

The contrastive capabilities of the encoder should be explored in the future as well, for example as a metric to train generative models.

% =================
%  Acknowledgement
% =================
\section{Acknowledgement}
The research leading to these results is funded by the German Federal Ministry for Economic Affairs and Climate Action" within the project “Verifikations- und Validierungsmethoden automatisierter Fahrzeuge im urbanen Umfeld". The authors would like to thank the consortium for the successful cooperation.

\bibliographystyle{IEEEtran}
\bibliography{references}

\end{document}